\documentclass[10pt,journal,cspaper,compsoc]{IEEEtran}
\usepackage{color}
\usepackage[pagebackref=true,breaklinks=true,letterpaper=true,colorlinks,bookmarks=false]{hyperref}

\ifCLASSOPTIONcompsoc

\else
   \usepackage{cite}
\fi

\ifCLASSINFOpdf
   \usepackage[pdftex]{graphicx}

\else
  \fi

\usepackage[cmex10]{amsmath}

\begin{document} 
%
\title{3D Interest Point Detection via Discriminative Learning \\}

\author{Leizer~Teran~and~Philippos~Mordohai~
\IEEEcompsocitemizethanks{\IEEEcompsocthanksitem L.Teran. Email: lzrteran@gmail.com 
\IEEEcompsocthanksitem P. Mordohai is with the Department
of Computer Science, Stevens Institute of Technology, Hoboken,
NJ, 07030. \protect\\
E-mail: mordohai@cs.stevens.edu}
\thanks{}}

\IEEEcompsoctitleabstractindextext{%
\begin{abstract} 
The task of detecting the interest points in 3D meshes has typically been handled by geometric methods. These methods, while greatly describing human preference, can be ill-equipped for handling the variety and subjectivity in human responses. Different tasks have different requirements for interest point detection; some tasks may necessitate high precision while other tasks may require high recall. Sometimes points with high curvature may be desirable, while in other cases high curvature may be an indication of noise. Geometric methods lack the required flexibility to adapt to such changes. As a consequence, interest point detection seems to be well suited for machine learning methods that can be trained to match the criteria applied on the annotated training data. In this paper, we formulate interest point detection as a supervised binary classification problem using a random forest as our classifier. Among other challenges, we are faced with an imbalanced learning problem due to the substantial difference in the priors between interest and non-interest points. We address this by re-sampling the training set. We validate the accuracy of our method and compare our results to those of five state of the art methods on a new, standard benchmark.
\end{abstract}

}

\maketitle

\IEEEdisplaynotcompsoctitleabstractindextext

%
\IEEEpeerreviewmaketitle

\section{Introduction} 

\begin{figure}[b!!]
\vspace{-10pt}
\begin{center}
\begin{tabular}{cc}
\includegraphics[width=.4\columnwidth]{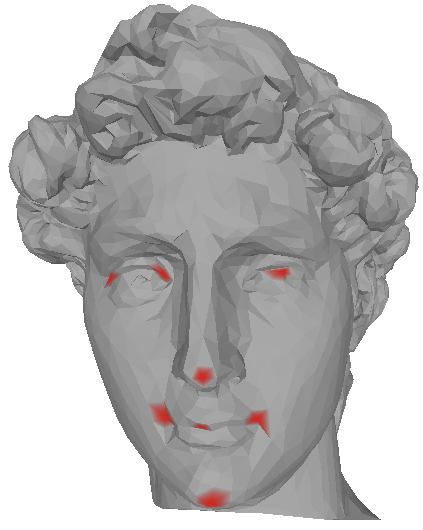} &
\includegraphics[width=.4\columnwidth]{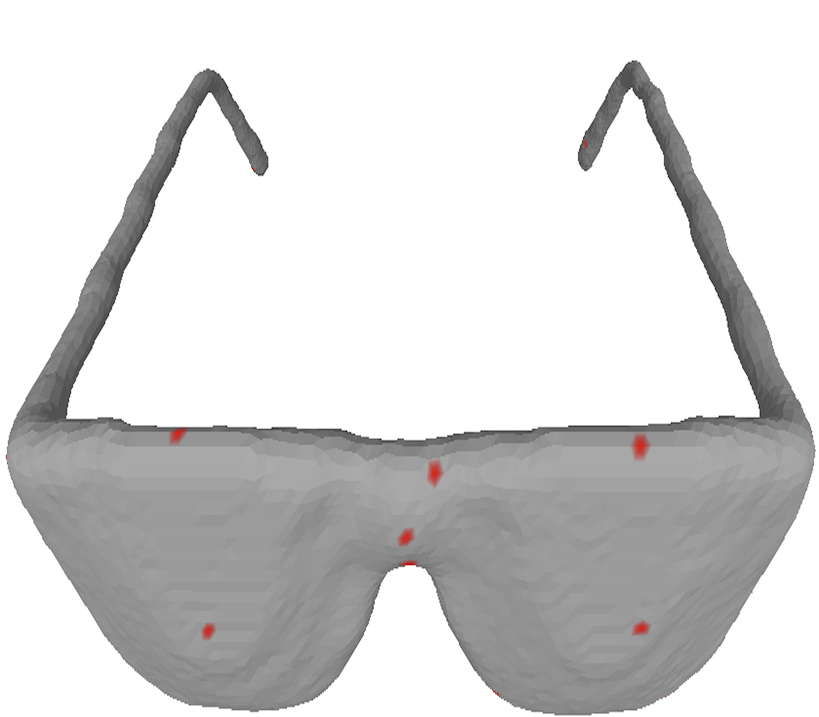} \\
\end{tabular}
\end{center}
\vspace{-10pt}
\caption{David and glasses ground truth points}\label{fig:intro1}
\end{figure}

\IEEEPARstart{T}{he}  identification of important points in images has been a long standing problem in computer vision. Once detected, these important, or interest, points are encoded in one of many invariant representations, such as SIFT \cite{lowe04}, and are used within a multitude of applications such as  image registration, retrieval, object tracking and structure from motion. Note that Lowe \cite{lowe04} presents techniques for detecting and describing interest points, but one can use a different detector and then apply the SIFT descriptor. A similar two-stage approach can be applied to 3D data. However, due to concerns about the reliability of interest point detectors in 3D, in many cases descriptors are computed at uniformly sampled locations of the 3D model \cite{johnson99,frome04,rusu09}. The reliability of 3D interest point detectors was recently studied by Dutagaci et al. \cite{dutagacibenchmark} who created a benchmark (Fig. \ref{fig:intro1}) and evaluated several state of the art methods. In this paper, we go beyond pure geometry for 3D interest point detection by learning to detect such points from a corpus of annotated data. Note that descriptor computation is out of scope here.

One of the main difficulties in predicting interest points lies in the discrepancy between quantified importance and perceived importance. Many methods assume that quantitatively important points, usually found by optimization of a function, correspond to perceptually important points. This assumption works well for vertices that are co-located with sharp changes in the model, e.g. corners. For smoother regions and perceptually ambiguous points, the previous assumption along with a multi-scale approach have been met with varying success. Another layer of difficulty arises when semantic ambiguity is considered. This is due to varying, task-specific requirements and, in the case of our data, due to subjectivity of the annotators.

A successful algorithm should be invariant to rotation, translation and scaling, capture points that are appealing to a large consensus of people and have enough flexibility to deal with ambiguity and subjectivity. One approach to capturing subjectivity, which has not been fully explored in the 3D shape analysis literature, is discriminative learning that attempts to identify patterns associated with the annotators' preferences.

\begin{figure}[b!!]
\begin{center}
\begin{tabular}{cc}
\includegraphics[width=.4\columnwidth]{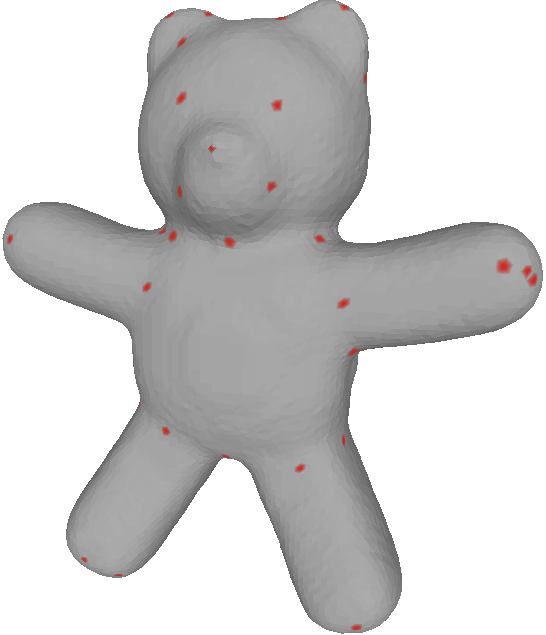} &
\includegraphics[width=.4\columnwidth]{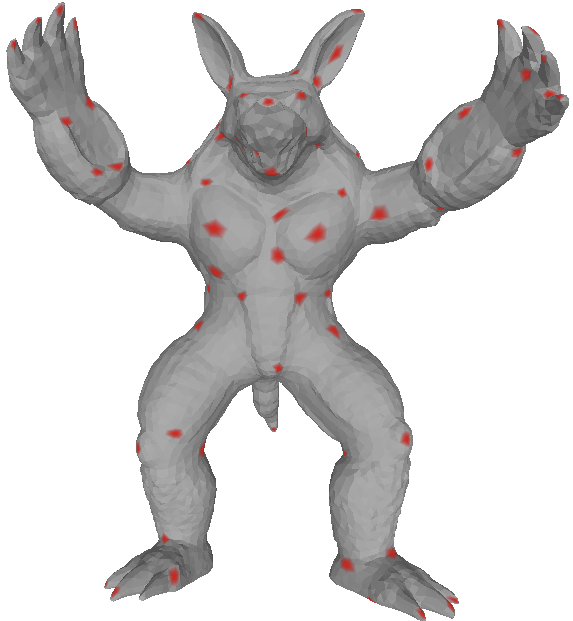} \\
\end{tabular}
\end{center}
\vspace{-10pt}
\caption{Teddy bear and armadillo ground truth points}\label{fig:intro2}
\end{figure}

In this paper, we formulate 3D interest point detection as a binary classification problem. We use several geometric detectors to produce attributes which are the inputs to a learning algorithm that gives competitive performance against five state of the art methods on the benchmark of Dutagaci et al. \cite{dutagacibenchmark}. The peculiarity of this benchmark is that the ground truth interest points were selected by non-expert users who were asked to click on the models. As a result, points with widely varying geometric properties are selected in each case. For example, as seen in Fig. \ref{fig:intro1}, the high curvature of David's hair does not attract the attention of the annotators. In fact in many cases, such as the teddy bear in Fig. \ref{fig:intro2}, the interest points lie on smooth hemi-spheres. Subjectivity and semantics play a large role in feature selection in this dataset posing significant challenges to geometric methods. Section \ref{subsec:results} shows that our approach is able to cope with this variability much better.

\section{Related Work}

In a recent survey, Dutagaci et al \cite{dutagacibenchmark} introduced a benchmark and an evaluation methodology for algorithms designed to predict interest points in 3D. The benchmark comprises 43 triangular meshes and the associated paper evaluated the performance of six algorithms \cite{meshsaliency,sdcorners,salientpoints,hks,3dharris11,3dsift} in interest point detection. Since we also use this benchmark, here, we focus our attention to these six methods. Other relevant methods include \cite{gelfand05,gal06,matei06,akagunduz07,shilane07,zhong09,knopp10}. We refer readers to recent surveys \cite{mian10,salti11,dutagacibenchmark,lian13,YuCipolla13} for more details.

\subsection*{Mesh Saliency.} Lee et al. \cite{meshsaliency} address interest point detection through the use of local curvature estimates coupled with a center surround scheme at multiple scales. The total saliency of a vertex is defined as the sum of Difference of Gaussian (DoG) operators over all scales.

\subsection*{Scale Dependent Corners.} Novatnack and Nishino \cite{sdcorners}  measure the geometric scale variability of a 3D mesh on a 2D representation of the surface geometry given by its normal and distortion maps, which can be obtained by unwrapping the surface of the model onto a 2D plane. A geometric scale-space which encodes the evolution of the surface normals on the 3D model
while it is gradually smoothed is constructed and interest points are extracted as points with high curvature at multiple scales. 

\subsection*{Salient Points.} Castellani et al. \cite{salientpoints} also adopt a multi-scale approach. DoG filters are applied to vertex coordinates to compute a displacement vector of each vertex at every scale. The displacement vectors are, then, projected onto the normals of the vertices producing a ``scale map'' for each scale. Interest points are extracted among the local maxima of the scale maps.

\subsection*{Heat Kernel Signature.} Sun et al. \cite{hks} apply the Laplace-Beltrami operator over the mesh to obtain its Heat Kernel Signature (HKS). The HKS captures neighborhood structure properties which are manifested during the heat diffusion process on the surface model and which are invariant to isometric transformations. The local maxima of the HKS are selected as the interest points of the model.

\subsection*{3D Harris.} Sipiran and Bustos \cite{3dharris11} generalized the Harris and Stephens corner detector \cite{Harris88} to 3D. The computation is now performed on the rings of a vertex, which play the role of neighboring pixels. A quadratic surface is fitted to the points around each vertex. This enables the computation of a filter similar to the Harris operator, the maximal responses of which are selected as interest points.

\subsection*{3D SIFT.} Godil and Wagan \cite{3dsift} initially convert the mesh model into a voxel representation. Then, 3D Gaussian filters are applied to the voxel model at various scales as in the standard SIFT algorithm. DoG filters are used to compute the difference between the original model and the model at a particular scale and their extrema are taken as candidate interest points. The final set of interest points are those that also lie on the surface of the 3D object.

\section{Attributes and Learning}\label{sec:learning} 

In this section, we focus on the geometric attributes that are used as the inputs to our classifier. The attributes capture characteristics that intuitively should help discriminate between interest and regular points, such as curvature, saliency compared to neighboring vertices, etc. Since interest points may become salient at different scales, we capture information at multiple scales by applying Difference of Gaussian (DoG) filters on the attributes \cite{meshsaliency,salientpoints}.
All attributes are invariant to rotation, translation and scale. The latter is achieved by normalizing the lengths in each mesh by its diameter.

First, we present the attributes that serve as the basic building blocks for all the others. The motivation behind this basic set is to create descriptors that capture the local geometric properties and context of a given vertex. We attempt to quantify the basic properties of every vertex, $v$, through the following 10 attributes, the first 7 of which use the 100 nearest Euclidean neighbors, denoted as $\nu(v)$.

\subsection{Basic Attributes}
The first 5 attributes involve the scatter matrix about a vertex $v$:
\begin{align*}
S(v)=\sum_{x \in \nu(v) }\exp^{-\frac{||{\bf x}-{\bf v}||^2}{\tau^2/2}}\frac{({\bf x}-{\bf v})({\bf x}-{\bf v})^T}{||{\bf x}-{\bf v}||^2},
\end{align*}
with ${\bf x}$ being the coordinates of vertex $x$ and ${\bf v}$ being the coordinates of vertex $v$. The parameter $\tau$, was taken to be the radius of $\nu(v)$. The contribution of each neighbor is weighted according to its proximity with $v$ to limit the influence of neighboring points that may be located across discontinuities.

Attributes 1 to 3 are ratios of eigenvalues of $S(v)$:
\begin{align*}
F_1(v)=&\lambda_{1,v}/\lambda_{2,v}\\
F_2(v)=&\lambda_{1,v}/\lambda_{3,v}\\
F_3(v)=&\lambda_{2,v}/\lambda_{3,v},
\end{align*}
where $\lambda_{i,v}$ is the $i$-th largest eigenvalue of $S(v)$. These capture properties of the surface, such as planarity in which case the $S(v)$ is almost rank deficient, or corners that have three eigenvalues of similar magnitude.

We look at the differences between the eigenvalues for the next two attributes, as follows:
\begin{align*}
F_4(v)=&\lambda_{2,v}-\lambda_{1,v}\\
F_5(v)=&\lambda_{3,v}-\lambda_{2,v}.
\end{align*}

Attribute 6 is the vertex density about the point, whereas attribute 7 is the average inner product of vertex $v$'s normal with the normals of its 100 nearest neighbors:
\begin{align*}
F_ 6(v)&=\frac{100}{\text{Vol}(\nu(v))}\\
F_ 7(v)&=\frac{\sum_{x \in \nu(v) }{\bf n}(v)\cdot {\bf n}(x)}{100}
\end{align*}
Where ${\bf n}(x)$ is the vertex normal of vertex $x$.

Attributes 8 and 9 are the principal curvatures at vertex $v$ and the 10$^{th}$ attribute is its Gaussian curvature.
\begin{align*}
F_ 8(v)&=\kappa_1(v)\\
F_ 9(v)&=\kappa_2(v) \\
F_{10}(v)&=\kappa_1(v)\kappa_2(v)
\end{align*}
To compute the principal curvatures, we look at the one-ring of the current vertex. Then, directional curvatures along the edges are approximated and used to compute the tensor of curvature \cite{taubin95}.

\subsection{DoG Attributes}
The basic attributes described above are generally not enough to detect all interest points, since they may become salient at different scales. Inspired by the success of the multi-scale approach adopted by other algorithms \cite{meshsaliency,salientpoints}, we compute a set of DoG attributes that are functions of the basic attributes.

The DoG attribute computations were performed within Euclidean neighborhoods of radius $r$ centered at vertex $v$, which will be referred to as $N(v,r)$. We found the Gaussian weighted average within the $\delta,2\delta,4\delta$ and $6 \delta$ neighborhoods of vertex $v$, for each basic attribute. We use 0.3\% of the model diameter as $\delta$. In addition, we compute the Gaussian weighted  neighborhood averages of the mean curvature for vertex $v$, which was not included in the set of basic attributes because it is a linear combination of the principal curvatures.

Attributes 11 through 20 are the DoGs between the $\delta $ and $2\delta$ neighborhoods at each vertex for each basic attribute. 
\begin{align*}
G_{\delta,j}(v)&=\frac{\sum_{x \in N(v,\delta)}F_{j}(x)\exp^{-||x-v||^2/(2\delta^2)}}{\sum_{x \in N(v,\delta)} \exp^{-||x-v||^2/(2\delta^2)} } \\
F_{j+10}(v)&=|G_{2\delta,j}(v)-G_{\delta,j}(v)|    \text{     }, j \in \{1...10\}.
\end{align*}

Attribute 21 is the DoG of the mean curvature for the $\delta$ and $2\delta$ neighborhoods:
\begin{align*}
\mu_{\delta}(v)&=\frac{\sum_{x \in N(v,\delta)}C(x)\exp^{-||x-v||^2/(2\delta^2)}}{\sum_{x \in N(v,\delta)} \exp^{-||x-v||^2/(2\delta^2)} }\\
F_{21}(v)&=|\mu_{2\delta}-\mu_{\delta}|
\end{align*}
with $C(x)$ being the mean curvature at vertex $x$.

Attributes 22 through 31 are DoGs for the basic attributes but use the $2\delta$ and $4\delta$ neighborhoods instead.
The 32$^{nd}$ attribute is the mean curvature DoG for the $2\delta$ and $4\delta$ neighborhoods.
Finally, we look at the DoGs for the $4\delta$ and $6\delta$ neighborhoods to define the 33$^{rd}$ through 43$^{rd}$ attributes in the same way.

To summarize, each vertex has a total of 43 attributes, denoted by $F_i(v)$ with $i \in {1,...,43}$. The attribute set for each vertex can be broken down into a set of basic attributes  $\{ F_1(v)...F_{10}(v)\}$  and a set of DoG attributes  $\{ F_{11}(v)...F_{43}(v)\}$ that are functions of the basic attributes at varying scales.  With the inputs to our classifier in place, we now shift our attention to the random forest.

\subsection{Random Forest}\label{sec: forest}

Random forest classifiers are ensembles of classification and regression trees that have gained popularity due to their high accuracy and ability to generalize \cite{breiman2001,criminisiforest}. The key idea during training is to generate decisions trees that partition the attribute space in a way that separates the training data according to their labels, interest and non-interest points in our case. In the training stage, a new training set is created for each tree by sampling with replacement (bootstrapping) from the original training set. Each node performs randomly generated tests on random subsets of the full attribute set. The attribute and threshold value that optimize a function of the input samples is selected and the data are divided to the node's children. The Gini gain and the information gain are standard functions used for the selection. We use the former in our analysis.

Once the forest has been trained, the test set vertices are fed to each trained tree in the forest. The current vertex is run down each tree and decisions are made at every node based on the optimal splits computed during training. This process continues until the terminal node is reached and a decision is made about the current vertex's class label. The class label that is output by the majority of the trees in the forest is assigned to the vertex. In addition to outputting the class label for a given input, the random forest can output class membership probabilities by averaging the class probabilities given by each tree. These probabilities correspond to the class proportions of the terminal node the test sample falls in. We use Scikit-Learn \cite{scikit-learn} to implement the random forests.

The performance of the forest is controlled by its parameters: the depth of each tree, the number of attributes at each node and the the number of trees in the forest. Following Breiman's recommendation  \cite{breiman2001}, we do not prune the trees.  In general, the full attribute set is not used while sampling at each node. This is done to keep the trees in the forest as uncorrelated as possible. Following common practice, we use $\sqrt{p}$ attributes at each node with $p$ being the total number of attributes.

\subsection{Imbalanced Classes} 

Imbalanced classes present a challenge for most classifiers. Poor predictive performance arises because the standard implementation aims to reduce the overall error rate. As a consequence, the random forest can afford to misclassify almost all the minority class examples and still achieve a very low error rate. To make matters worse, the minority class may not even be selected during bootstrapping and therefore may be missing, entirely or essentially, during the training process. For our problem, the ratio of interest to non interest points can range anywhere from 1:100  to even 1:240 within our training data set.

There are a few strategies to deal with the misrepresentation of classes. We chose a technique proposed by Chen et al \cite{imbalancedforest} where the dominant class is down-sampled, while the minority class is over-sampled. For a set of labelled vertices, we randomly select $n$ interest points, where $n$ is one half of the total interest points. During training each tree is given a balanced set of vertices that have a ratio of, $k$ non interest vertices to $n$ interest vertices, where $k \geq n$. The parameter $k$ and the bootstrap ratio will be discussed in Section \ref{sec:exp}.

\section{Ground Truth and Experiments}\label{sec:exp}
In this section, we describe the data we used, the experimental setup and our results.

 \subsection{Ground Truth}
Dutagaci et al. \cite{dutagacibenchmark} used a web-based application to collect user clicks on 43 mesh models. These models were organized in two overlapping data sets, Data Set A and Data Set B, consisting of 24 and 43 triangular mesh models respectively. Through the web-based application, a user was shown the models from a data set one at a time and was allowed to freely click on them. Data Set A was annotated by 23 human subjects while Data Set B was annotated by 16 human subjects. The positions of the individual user clicks showed some variability as well as some consensus. The variability may be due to imprecise clicking or the subjective nature of interest points.

In order to determine user consensus and remove outliers, the authors considered two criteria while constructing each set's ground truth. The first is the radius, $\sigma d_M$, of an interest region and the second is the number of users $n$ that clicked within the region. The radius of the interest region is model specific with $d_M$ denoting the diameter of the model and $\sigma$ is a parameter in [0.01, 0.02, 0.03, ... , 0.1]. Individual user clicks are clustered together if their geodesic distances are less than $2\sigma d_M$. If the number of clicks in a cluster is less than $n$ that cluster is ignored. If not, the point that minimizes the sum of geodesic distances to the other points is chosen as the representative of the cluster and included in the ground truth. In case the distance between cluster representatives is less than $2\sigma d_M$, the clusters are merged and the representative with the highest number of cluster points is chosen as the final representative.

\begin{figure}[b]
\begin{center}
\begin{tabular}{cc}
\includegraphics[width=.18\textwidth]{figs/03_2_teddy} &
\includegraphics[width=.18\textwidth]{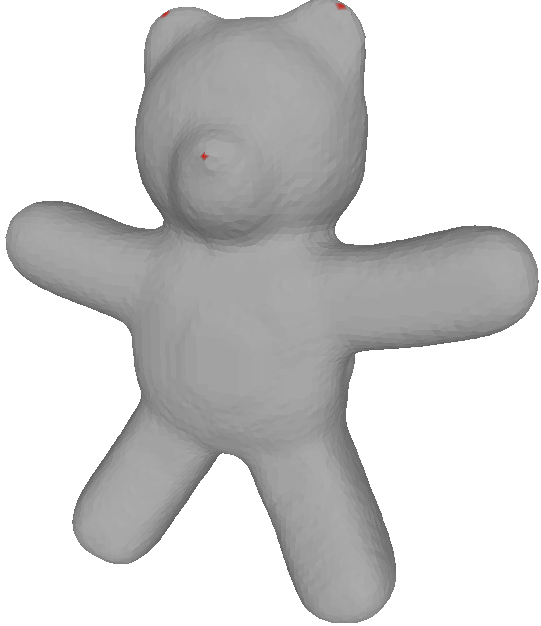} \\
(a) $\sigma=0.03 , n=2$ & (b) $\sigma=0.03 , n=8$  \\
\includegraphics[width=.18\textwidth]{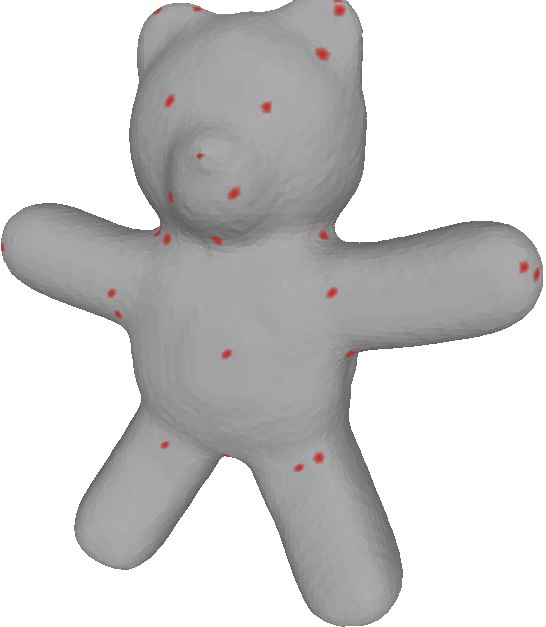} &
\includegraphics[width=.18\textwidth]{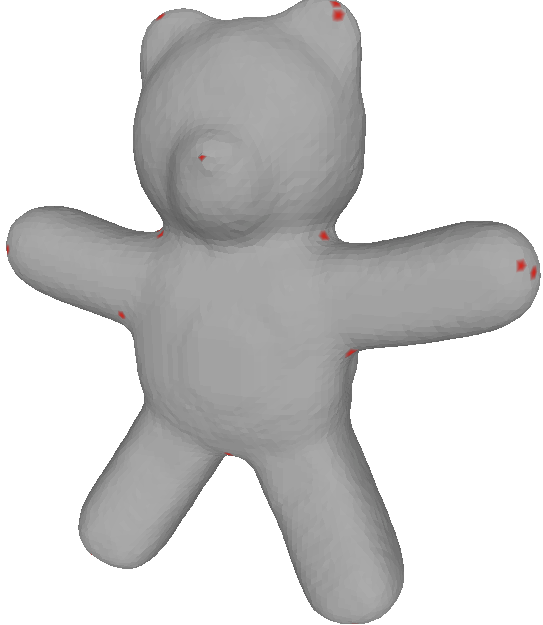} \\
(c) $\sigma=0.05 , n=2$ & (d) $\sigma=0.05 , n=8$  \\
\end{tabular}
\end{center}
\vspace{-8pt}
\caption{Ground truth points for the teddy bear for four $\sigma/n$ combinations.}\label{fig:teddy_combos}
\vspace{-8pt}
\end{figure}

The parameters, $\sigma$ and $n$, affect the number of ground truth points. For a fixed $\sigma$, fewer ground truth points are observed as $n$ increases since a higher consensus among users is needed. With small $\sigma$ and large $n$, there tends to be better localization around the points and typically fewer ground truth points are seen. As can be seen in Fig. \ref{fig:teddy_combos},  when $\sigma$ increases more ground truth points are observed. This trend continues until the clusters are large and close enough to be merged with adjacent clusters. Consequently, decreasing the overall number of ground truth points for large $\sigma$. 

 \subsection{Experiments}
We use the human generated ground truth to compare the random forest with the five following methods: Mesh Saliency \cite{meshsaliency}, Salient Points \cite{salientpoints}, Heat Kernel Signature \cite{hks}, 3D Harris \cite{3dharris11} and Scale Dependent Corners \cite{sdcorners}, across Data Sets A and B. There are a few models within these data sets that are not watertight and therefore do not allow volumetric representations. As a consequence, the output of 3D SIFT for these models was not provided at the benchmark website. This led to its exclusion from our analysis, but partial results can be seen in \cite{dutagacibenchmark}. (3D SIFT is not among the top performing methods on the benchmark.)

Sets A and B are treated as distinct experiments where we measure the effects of user consensus, $n$, and ground truth localization, $\sigma$, on algorithmic performance. Adhering to the protocol \cite{dutagacibenchmark}, we adopt the following $\sigma/n$  combinations for Set A: 0.03/2, 0.03/11, 0.05/2 and 0.05/11. For Set B we choose the following combinations: 0.03/2, 0.03/8, 0.05/2 and 0.05/8. In other words, for a given mesh model $M$, we assess how well the algorithms perform in detecting ground truth points agreed upon by at least two subjects or by at most one half of the subjects labeling each set.

\subsection{Training and Predicting}

For all experiments, we apply three-fold cross validation for both Sets A and B. Specifically, Data Set B was partitioned into three disjoint sets consisting of 14 models each (igea was removed since igea and bust2 are duplicates). Data Set A was split into three disjoint sets consisting of eight models each. Once the splits are established for a given set, we train the random forest, using the attributes from Section \ref{sec:learning} as the classifier input, on the first two folds and predict the interest points of the last fold, then the folds are rotated between the test and training sets.

Every vertex of every model in the training set of a given fold has a set of labels. Because the ground truth points vary by $\sigma$ and $n$, we have to make a compromise between the number of ground truth points available and high consensus among the annotators. For Data Set B, the representative of clusters $\sigma/n$, $n \in [11...22]$, are placed in the positive class for all values of $\sigma$ provided. All other vertices are placed in the negative class. Likewise, for Data Set A we place the representative of clusters  $\sigma/n$, $n \in [8...15]$, in the positive class for every available value of $\sigma$. The remaining vertices are considered members of the negative class.

Our classifier has three main parameters. The first is the bootstrap ratio of interest vertices to non-interest vertices that is used to balance the training set. The second is the number of attributes sampled during the node splitting process and the third is the number of trees. We use a 1:1 ratio of interest vertices to non-interest vertices while sampling and a random sample of 5 attributes at every node for all 100 unpruned trees in the forest. These parameters are found via cross validation.

In general, the forest returns a large number of interest points, as nearby vertices that have similar attributes form clusters. To address this, non-max suppression is performed on the test set vertices that are chosen by the forest. First, a set of candidate vertices is created by keeping the test vertices with interest class probability (Sec. \ref{sec: forest}) larger than 0.5. Then, the final set of interest vertices is created by keeping those vertices with an interest class probability larger than their neighbors. This suppression is done within the $c\psi$ Euclidean neighborhood of the chosen vertices, where $c$ is found to be 5 for Set A and 2 for Set B via cross validation. $\psi$ is the average minimum geodesic distance between ground truth vertices of the training set.

\subsection{Evaluation Criteria}\label{subsec:evals}

Dutagaci et al. \cite{dutagacibenchmark} evaluated the performance of the algorithms based on the following definitions and criteria.
Let $A_M$ be the set of vertices selected by the algorithm for a given mesh model. A ground truth point, $g \in G$, is correctly identified if there exists $a \in A_M$ such that the geodesic distance between them is less than some error tolerance ($\epsilon$) and that no other point in $G$ is closer to $a$. Or in other words, $g_0$ is correctly identified by the algorithm if: $g_0=\text{argmin}_{g \in G}(d(a,g)\leq \epsilon)$ for some $a \in A_M$.  The following error tolerances are used: $\epsilon=rd_M$ with $r \in [0,0.1,0.2...0.12]$.

This definition allows each correctly detected point in the ground truth set to be in correspondence with a unique $a \in A$. The number of false positives is then $N_A-N_C$ and the false negatives are $N_G-N_C$ where $N_A$, $N_C$ and $N_G$ are the number of algorithm selected points, correctly identified points and number of ground truth points respectively. The geodesic distances were computed using publicly available software \cite{Surazhskygeodesics}.

In addition to the evaluation criteria proposed by the authors of the benchmark, we also adopt the Intersection Over Union (IOU) criterion which has been used to evaluate object detection in images \cite{Everingham10}. The IOU for a given mesh model $M$ is defined as:
 \begin{align*}
 \text{IOU$_M$}=\frac{\text{TP}}{\text{FP}+\text{TP}+\text{FN}}\\
 \end{align*}

The definition for the IOU given above is for a mesh model. To find the IOU score over a set of models, as is done in the experiments, a running total of the false positives, false negatives and true positives were kept over all vertices in the data set and then used to compute a set-wise IOU. We compute the set-wise IOU score for the $\sigma/n$ combinations in the experiments section. For a given $\sigma/n$ combination the IOU scores are found for values of  $r \in  [0,0.1,0.2...0.12]$.

\subsection{Results}\label{subsec:results}
\begin{figure*}[b!!]
\begin{center}
\begin{tabular}{ccc}
\includegraphics[width=.62\columnwidth]{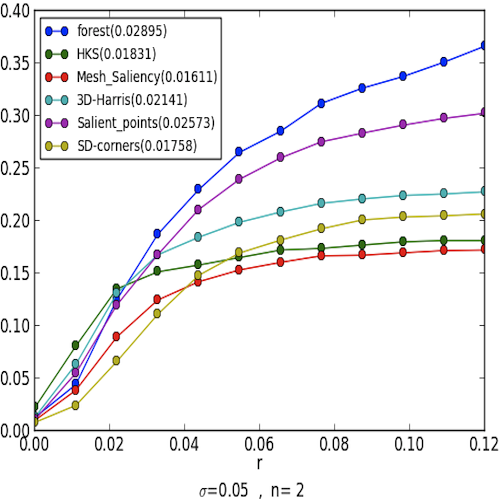}&\includegraphics[width=.62\columnwidth]{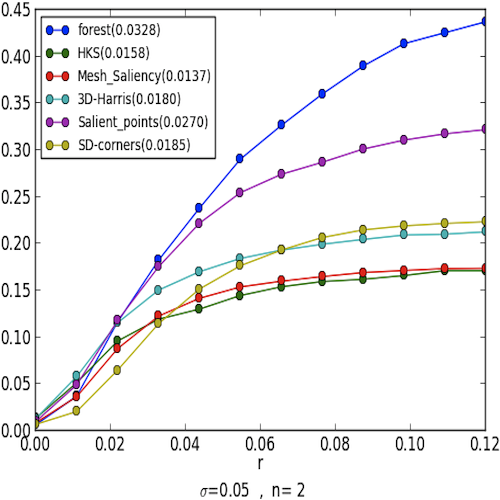}& \includegraphics[width=.62\columnwidth]{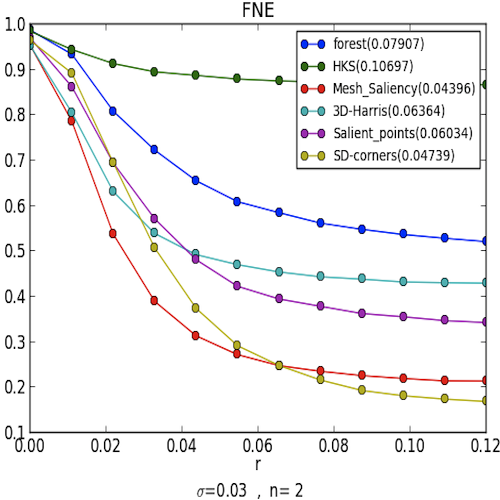} \\
\includegraphics[width=.62\columnwidth]{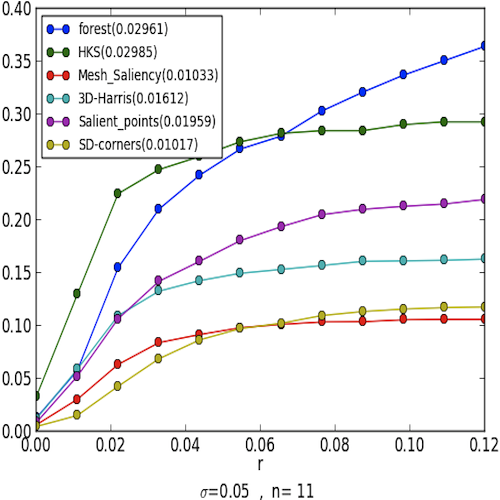}&\includegraphics[width=.62\columnwidth]{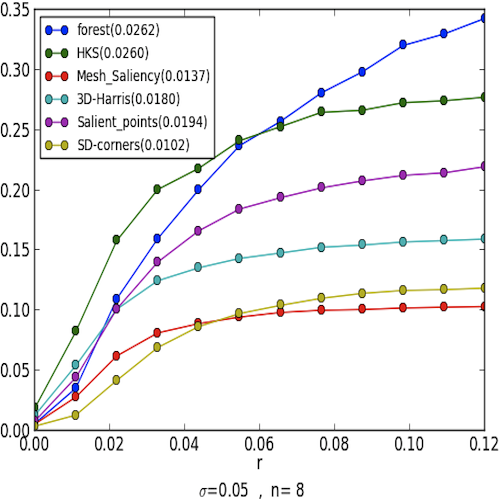} &\includegraphics[width=.62\columnwidth]{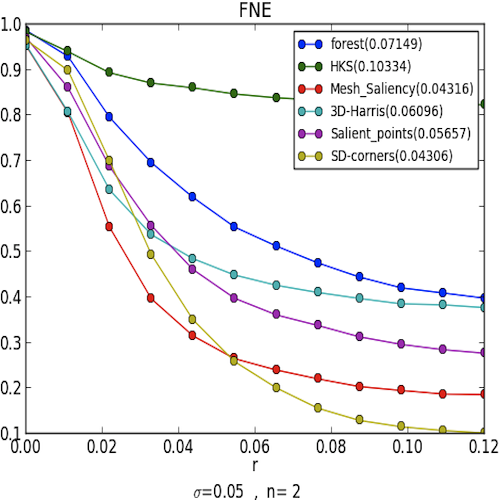} \\
\includegraphics[width=.62\columnwidth]{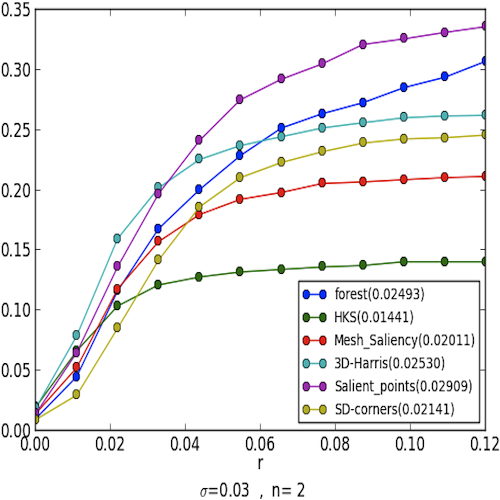}&\includegraphics[width=.62\columnwidth]{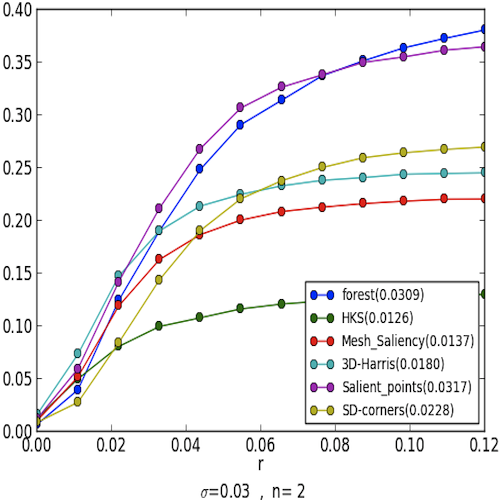} &\includegraphics[width=.62\columnwidth]{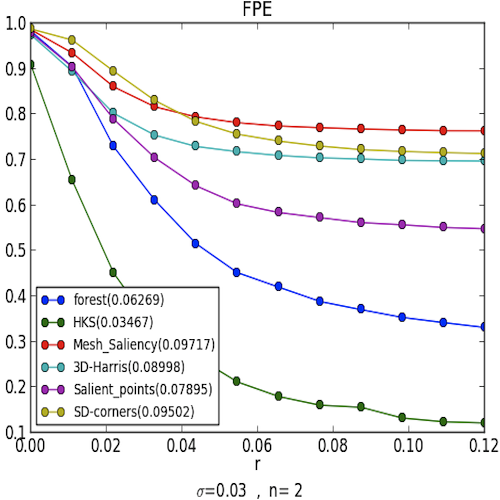} \\
\includegraphics[width=.62\columnwidth]{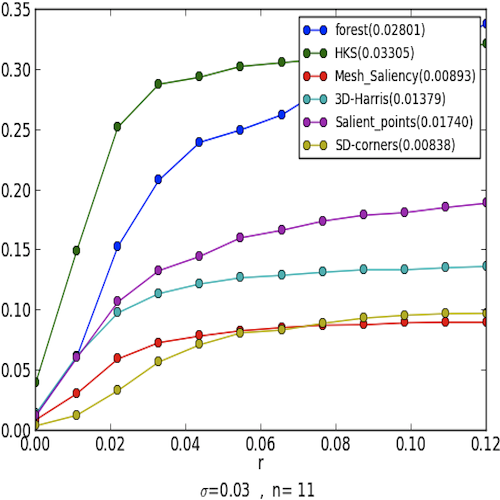}&\includegraphics[width=.62\columnwidth]{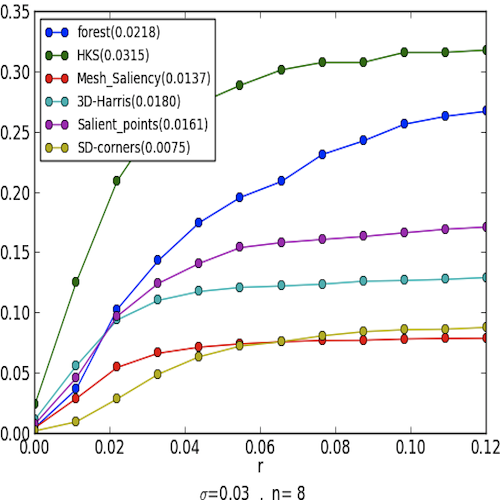} &\includegraphics[width=.62\columnwidth]{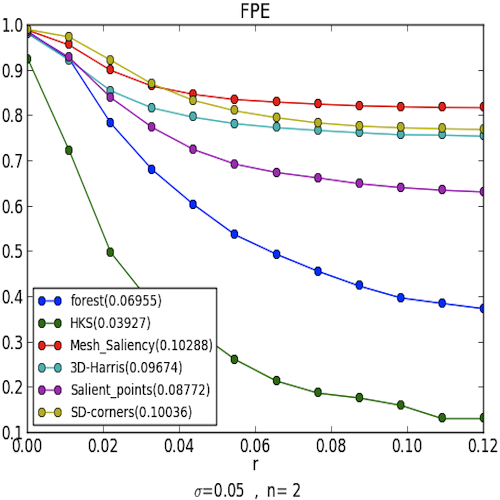} \\
a) Set A  &b) Set B &c) FNE/FPE
\end{tabular}
\caption{Column 1 shows the IOU curves for Set A at various $\sigma/n$ pairs. Column 2 contains the IOU curves for Set B and Column 3 has the FNE and FPE rates for SetB with $\sigma=0.03,n=2$ and $\sigma=0.05,n=2$. The parameter $r$ is mentioned in Sec. \ref{subsec:evals}. The AUC for each method is provided in the legend.}\label{fig: IOU_setB_combs}
\end{center}
\vspace{-10pt}
\end{figure*}

In this section, we present the results of the algorithm evaluations for Data Set A and Data Set B under the IOU metric. Fig.  \ref{fig: IOU_setB_combs} shows the results of the set-wise IOU scores averaged over the three folds. The legends show the Area Under the Curve (AUC) for the different values of the radius $r$. In addition, we compare the methods at $\sigma=0.03/n=2$ and $\sigma=0.05/n=2$ using the evaluation criteria proposed by \cite{dutagacibenchmark} in the last column of Fig. \ref{fig: IOU_setB_combs}. It is important to note that the points detected by the methods are constant when $\sigma$ and $n$ change.

\begin{figure}[h!!]
\begin{center}
\begin{tabular}{cc}
\includegraphics[width=.4\columnwidth]{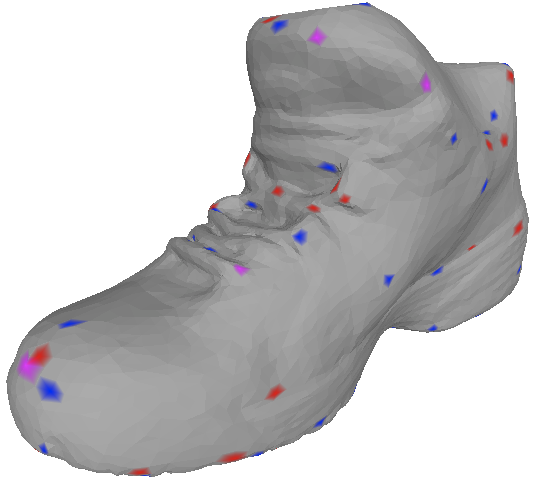} &
\includegraphics[width=.4\columnwidth]{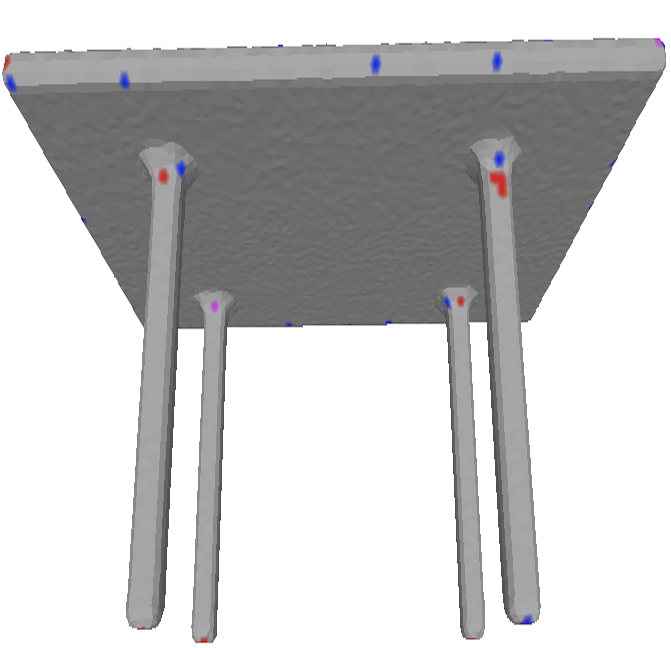} \\
a) Shoe: $\sigma=0.05,n=2$ &b) Table: $\sigma=0.03,n=2$\\
\includegraphics[width=.4\columnwidth]{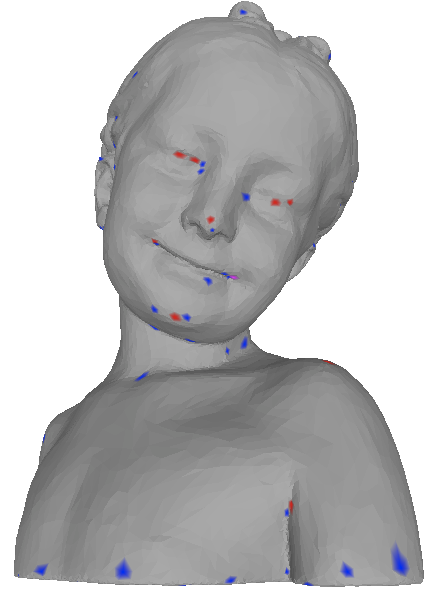} &
\includegraphics[width=.4\columnwidth]{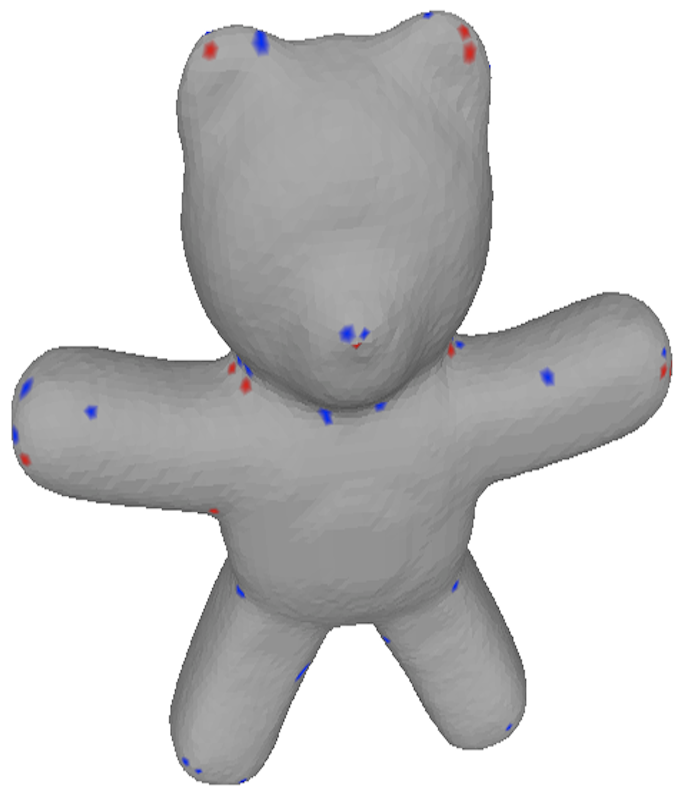} \\
c) Girl: $\sigma=0.03,n=8$ &d) Teddy: $\sigma=0.05,n=8$\\
\includegraphics[width=.4\columnwidth]{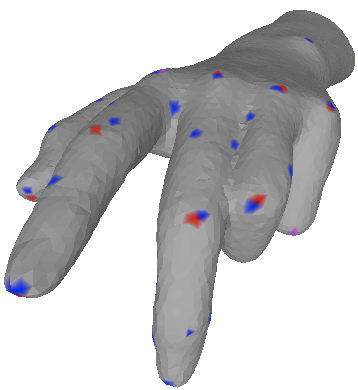} &
\includegraphics[width=.4\columnwidth]{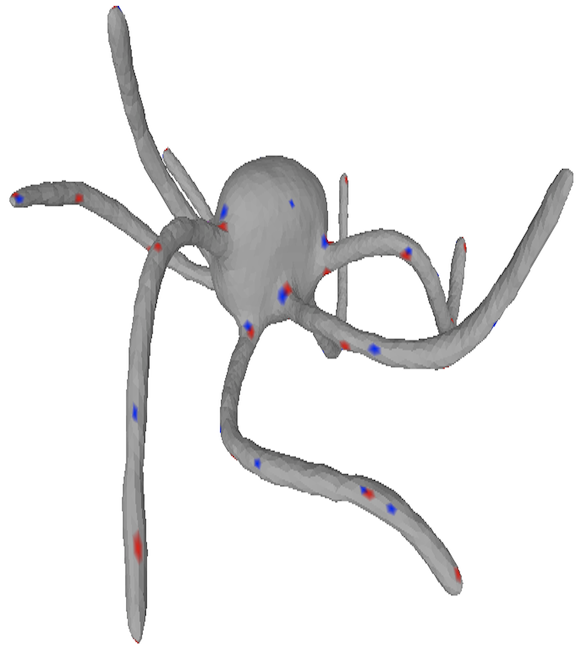} \\
e) Hand: $\sigma=0.03,n=8$ &f) Octopus: $\sigma=0.03,n=2$\\
\end{tabular}
\end{center}
\vspace{-6pt}
\caption{Random Forest predictions for six Set B models. Ground truth points for various $\sigma/n$ pairs in red. Random forest predictions in blue. Purple points are ground truth vertices that are predicted exactly.}\label{fig: forest_setB_combs}
\end{figure}

In Fig. \ref{fig: IOU_setB_combs} we see that the random forest is either first or second across Set B's $\sigma/n$ settings. We perform especially well when the localization is relaxed, as is seen when $\sigma=0.05$. One possible cause for this is that new ground truth points emerge when $\sigma$ increases for $n=8$. These new vertices are the representatives of the clusters formed by large user consensus, but that fall below the previous localization threshold. This is indicative of an ambiguous region within the model, as seen in the teddy bear's neck in Fig. \ref{fig: forest_setB_combs}. The forest captures these regions more effectively than the other methods.

Small values of $n$ result in large numbers of ground truth interest points, favoring aggressive methods, such as Salient Points. On the other hand, when $n$ is large, conservative methods are able to identify the truly salient points without producing too many false positives. In this sense, the HKS algorithm is an outlier compared to the rest of the algorithms, including ours. HKS can reliably detect a very small number of very salient points but is unable to detect even slightly more ambiguous points. The last column of Fig. \ref{fig: IOU_setB_combs} contains FPE and FNE curves for Set B with $n=2$. These curves reveal how aggressive or conservative the algorithms are. The random forest performs well according to all criteria over the parameter range. 

Set A is expected to be more challenging for a learning-based method, since the training set is smaller. Nevertheless, our algorithm is the top performing one according to IOU.

To reach an overall ranking, we average the IOU AUCs over all settings. This yields the following results in order of decreasing average AUC: random forest 0.02787; HKS 0.02390; Salient Points 0.02295; 3D Harris 0.01915; SD Corners 0.01438; and Mesh Saliency 0.01387. Repeating the same operation for set B, we obtain: random forest 0.0279; Salient Points 0.0236; HKS 0.0215; 3D-Harris 0.0180; SD corners 0.0148; and Mesh Saliency 0.0137.

\section{Conclusions}
We presented a discriminative learning approach to 3D interest point detection that gives competitive performance over state of the art methods. Experiments on a new, publicly available benchmark demonstrate that our method handles variability in the ground truth, or the desirable output, more steadily than other methods.  This translates to an increased ability to cope with human subjectivity in these experiments, but it is equivalent to the ability to cope with different task-specific requirements imposed on the algorithm. 

\appendices
\
\ifCLASSOPTIONcompsoc
 \else
  \section*{Acknowledgment}
\fi

\ifCLASSOPTIONcaptionsoff
  \newpage
\fi

\bibliographystyle{IEEEtran}
\bibliography{interest_points}

\end{document}